\documentclass[sigconf]{acmart}
\AtBeginDocument{%
  }

\setcopyright{acmlicensed}

\copyrightyear{2026}
\acmYear{2026}
\setcopyright{cc}
\setcctype{by}
\acmConference[CIKM '26]{Proceedings of the 35th ACM International Conference on Information and Knowledge Management}{November 07--11, 2026}{Rome, Italy}
\acmBooktitle{Proceedings of the 35th ACM International Conference on Information and Knowledge Management (CIKM '26), November 07--11, 2026, Rome, Italy}
\acmDOI{10.1145/3799682.3840143}
\acmISBN{979-8-4007-2539-5/2026/11}

\usepackage{newtxmath}
\usepackage{amssymb}
\usepackage{amsmath}
\usepackage{multirow}
\usepackage[linesnumbered,ruled,vlined]{algorithm2e}
\usepackage{algpseudocode}
\usepackage{listings}
\usepackage{xcolor}
\usepackage{float}
\usepackage{url}
\usepackage{enumitem}
\usepackage{makecell}
\usepackage{graphicx}
\usepackage{fontawesome5}
\usepackage{balance}
\definecolor{mainblue}{RGB}{25, 25, 112}
\definecolor{accentred}{RGB}{178, 34, 34}
\definecolor{graytext}{RGB}{80, 80, 80}
\usepackage{tikz} 
\usetikzlibrary{positioning, calc, arrows.meta, backgrounds}
\newcommand{\mypara}[1]{\vspace{0.5em}\noindent\textbf{#1}}

\newif\ifcomments
\commentstrue  

\ifcomments
    \usepackage{xcolor}
    \newcommand{\zy}[1]{\textcolor{orange}{\textbf{[zhuyun: #1]}}}
    \newcommand{\liuyc}[1]{\textcolor{blue}{\textit{[liuyc: #1]}}}
    \newcommand{\luoyc}[1]{\textcolor{teal}{\textsf{[luoyc: #1]}}}
\else
    \newcommand{\zy}[1]{}
    \newcommand{\liuyc}[1]{}
    \newcommand{\luoyc}[1]{}
\fi

\begin{document}

\newcommand{\stitle}[1]{\vspace{0.4ex}\noindent{\textbf {#1}}}
\newcommand{\todo}[1]{{\color{red}{#1}}}

\newcommand{\Boruta}{$\mathsf{Boruta}$}
\newcommand{\Gealearning}{$\mathsf{GeaLearning}$}
\newcommand{\GeaLearning}{$\mathsf{GeaLearning}$}
\newcommand{\gealearning}{$\mathsf{GeaLearning}$}

\newcommand{\our}{GraphFAS\xspace}
\newcommand{\eg}{\textit{e.g.}}
\newcommand{\ie}{\textit{i.e.}}

\title{\our: A Distributed System for Automated Graph Feature Generation and Selection in Industrial Transaction Networks}


\author{Yice Luo}
\affiliation{%
  \institution{Ant Group}
  \city{Hangzhou}
  \country{China}
}

\author{Yun Zhu}
\affiliation{%
  \institution{Ant Group}
  \city{Hangzhou}
  \country{China}
}

\author{Xi Chen}
\affiliation{%
  \institution{Ant Group}
  \city{Hangzhou}
  \country{China}
}

\author{Yongchao Liu}
\authornote{Corresponding author.}
\affiliation{%
  \institution{Ant Group}
  \city{Hangzhou}
  \country{China}
}
\email{yongchao.ly@antgroup.com}

\author{Xintan Zeng}
\affiliation{%
  \institution{Ant Group}
  \city{Hangzhou}
  \country{China}
}

\author{Chengying Huan}
\affiliation{%
  \institution{Nanjing University}
  \city{Nanjing}
  \country{China}
}

\author{Kai Zhang}
\affiliation{%
  \institution{Ant Group}
  \city{Hangzhou}
  \country{China}
}

\author{Jinrui Zhang}
\affiliation{%
  \institution{Ant Group}
  \city{Hangzhou}
  \country{China}
}

\author{Juelu Zhang}
\affiliation{%
  \institution{Ant Group}
  \city{Hangzhou}
  \country{China}
}

\author{Jiajun Zheng}
\affiliation{%
  \institution{Ant Group}
  \city{Hangzhou}
  \country{China}
}

\thanks{$^\dagger$Corresponding authors; We thank Mingyao Li, Yuhang Chen, Yue Jin and Chuntao Hong for their help and contributions.}

\renewcommand{\shortauthors}{Yice Luo et al.}

\begin{abstract}
Industrial fraud detection often relies on costly expert-crafted features that overlook graph-structured relational signals, while GNNs often do not meet the interpretability and deployment requirements of financial risk control. We propose \our (\underline{G}raph \underline{F}eature \underline{A}utomated \underline{S}election), a distributed feature selection procedure based on Boruta that bridges this gap through: (1) a non-parametric graph feature generation module that constructs explicit, interpretable structural features via multi-hop subgraph extraction and multi-scale aggregation without learned parameters; and (2) an automated distributed feature selection algorithm extending Boruta with median-based aggregation across partitions to robustly identify informative features at scale with minimal domain expertise. Compared with end-to-end GNN pipelines, \our decouples feature aggregation from model training, enabling direct integration with tabular models and direct compatibility with TreeSHAP-based explanations. Deployed in Alipay, \our delivers order-of-magnitude improvements in engineering efficiency while showing strong performance against expert-driven and graph-learning baselines on large-scale graphs.
\end{abstract}

\ccsdesc[500]{Information systems~Data mining}
\keywords{Feature Selection, Graph Feature Generation, Interpretability, Distributed Graph Mining, Fraud Detection}

\maketitle
  
\definecolor{apipink}{rgb}{0.858, 0.188, 0.478}

\lstdefinestyle{apistyle}{
    aboveskip=0pt,
    belowskip=0pt,
    belowcaptionskip=1\baselineskip,
    breaklines=true,
    frame=none,
    numbers=left,
    xrightmargin=1em,
    framexrightmargin=1em,
    basicstyle=\footnotesize\ttfamily,
    keywordstyle=\bfseries\color{apipink},
    commentstyle=\itshape\color{green!40!black},
    identifierstyle=\color{black},
    backgroundcolor=\color{gray!10!white},
    linewidth=.97\columnwidth,
    numbersep=3pt,
}
\section{Introduction}

Fraud detection~\cite{fraud_survey,fraud1} in large-scale transaction networks is a critical task for financial platforms. On systems such as Alipay, effective detection mechanisms prevent financial losses amounting to millions of CNY daily. These transaction networks comprise hundreds of millions of users and massive edge volumes, where fraudulent activities exhibit organized collusion and complex interaction patterns inherently suited for graph-based analysis.

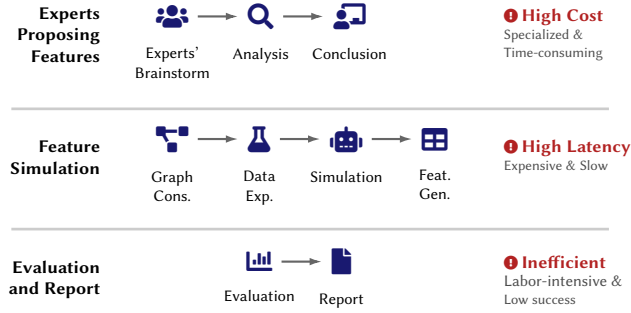
\begin{figure}[t]
\centering
\resizebox{\linewidth}{!}{
\begin{tikzpicture}[
    node distance=0.5cm and 0.8cm,
    font=\sffamily,
    stage_title/.style={
        text=black,
        font=\sffamily\bfseries\large,
        align=right,
        anchor=east
    },
    process_node/.style={text=mainblue, font=\huge, align=center, inner sep=5pt},
    every label/.style={text=black, font=\sffamily\normalsize, align=center, yshift=-2pt},
    pain_point/.style={
        text=accentred,
        font=\sffamily\small,
        align=left,
        anchor=west
    },
    myarrow/.style={-{Latex[length=2.5mm, width=1.5mm]}, draw=black!60, line width=1.2pt}
]
    \node[stage_title] (title1) {Experts\\Proposing\\Features};
    \node[process_node, right=0.8cm of title1, yshift=0.35cm, label=below:{Experts'\\Brainstorm}] (icon1_1) {\faUsers};
    \node[process_node, right=of icon1_1, label=below:{Analysis}] (icon1_2) {\faSearch};
    \node[process_node, right=of icon1_2, label=below:{Conclusion}] (icon1_3) {\faChalkboardTeacher};
    \draw[myarrow] (icon1_1) -- (icon1_2);
    \draw[myarrow] (icon1_2) -- (icon1_3);

    \node[stage_title, below=2.4cm of title1.east, anchor=east] (title2) {Feature\\Simulation};
    \node[process_node, right=0.8cm of title2, yshift=0.35cm, label=below:{Graph\\Cons.}] (icon2_1) {\faProjectDiagram};
    \node[process_node, right=of icon2_1, label=below:{Data\\Exp.}] (icon2_2) {\faFlask};
    \node[process_node, right=of icon2_2, label=below:{Simulation}] (icon2_3) {\faRobot};
    \node[process_node, right=of icon2_3, label=below:{Feat.\\Gen.}] (icon2_4) {\faTable};
    \draw[myarrow] (icon2_1) -- (icon2_2);
    \draw[myarrow] (icon2_2) -- (icon2_3);
    \draw[myarrow] (icon2_3) -- (icon2_4);

    \node[stage_title, below=2.4cm of title2.east, anchor=east] (title3) {Evaluation\\and Report};
    \node[process_node, label=below:{Evaluation}] (icon3_1) at ($(icon2_2 |- title3) + (0, 0.35cm)$) {\faChartBar};
    \node[process_node, right=of icon3_1, label=below:{Report}] (icon3_2) {\faFile};
    \draw[myarrow] (icon3_1) -- (icon3_2);

    \coordinate (pain_start_x) at ($(icon2_4.east) + (0.8cm, 0)$);
    \node[pain_point] (pain1) at (pain_start_x |- title1) {\faExclamationCircle\ \textbf{\large High Cost}\\\textcolor{black!70}{\small Specialized \&}\\\textcolor{black!70}{Time-consuming}};
    \node[pain_point] (pain2) at (pain_start_x |- title2) {\faExclamationCircle\ \textbf{\large High Latency}\\\textcolor{black!70}{\small Expensive \& Slow}};
    \node[pain_point] (pain3) at (pain_start_x |- title3) {\faExclamationCircle\ \textbf{\large Inefficient}\\\textcolor{black!70}{\normalsize Labor-intensive \&} \\\textcolor{black!70}{Low success}};

    \coordinate (left_limit) at (title1.west);
    \coordinate (right_limit) at (pain1.east);
    \coordinate (line1_y) at ($(title2.north) + (0, 0.5cm)$);
    \coordinate (line2_y) at ($(title3.north) + (0, 0.5cm)$);
    \draw[black!20, line width=1.5pt] (left_limit |- line1_y) -- (right_limit |- line1_y);
    \draw[black!20, line width=1.5pt] (left_limit |- line2_y) -- (right_limit |- line2_y);
\end{tikzpicture}
}
\caption{Traditional feature engineering workflows. The manual process is high-cost, while simulation is computationally expensive.}
\label{fig:exp_gnn}
\end{figure}

Existing industrial solutions predominantly rely on expert-crafted features developed through labor-intensive manual processes. As illustrated in Figure~\ref{fig:exp_gnn}, graph feature engineering entails three stages: (1) expert-driven pattern analysis requiring specialized domain knowledge, (2) large-scale simulations over massive credit networks, and (3) iterative multi-dimensional evaluations for stability validation. This manual cycle often spans over a month with no guarantee of optimal outcomes. Moreover, conventional attribute-centric approaches may under-utilize the rich relational signals embedded in graph structures, leaving critical topological patterns undetected.

Graph neural networks (GNNs) capture structural dependencies through iterative message passing, achieving strong performance on various graph learning tasks. However, GNNs face fundamental limitations in financial risk control scenarios. First, their multi-layer transformations can make decision processes harder to interpret that conflict with regulatory requirements for model transparency. Second, the computational overhead of end-to-end training introduces significant latency that conflicts with low-latency feature production requirements on large-scale industrial networks. Third, GNN embeddings often lack explicit semantics that are easy to validate in risk analysis workflows, making them difficult to validate and deploy in production environments.

To overcome the limitations of both manual feature engineering and opaque GNN embeddings, recent research has explored automated graph feature construction. Frameworks such as TAG~\cite{TAG}, G2T-FM~\cite{G2T-FM}, and TabPFN-GN~\cite{TabPFN-GN} construct node representations from topology using predefined structural encoders. While these approaches reduce manual intervention, they introduce new challenges that motivate our work:

\begin{itemize}[leftmargin=*]
\item \textbf{Challenge 1: Lack of interpretability in graph learning models.}
GNNs generate latent embeddings that are opaque to domain experts and regulators. Financial risk control requires explicit, human-interpretable features that can be directly validated and audited. The first challenge is to design a feature generation mechanism that captures multi-hop relational dependencies while producing semantically meaningful, tabular-compatible features with native TreeSHAP explainability (Section~\ref{sec:method_generation}).

\item \textbf{Challenge 2: Scalability limitations of existing feature selection methods.}
Traditional feature selection algorithms such as Boruta~\cite{JSSv036i11} operate on standalone environments and lack distributed computation support. Industrial graphs with massive edge volumes require partition-based processing, but aggregating importance scores across partitions introduces instability under skewed class distributions. The second challenge is to design a distributed adaptation with robust aggregation mechanisms for efficient high-dimensional feature screening across partitioned datasets (Section~\ref{sec:method_selection}).

\sloppy
\item \textbf{Challenge 3: Inflexibility of automated graph feature pipelines.}
Existing automated methods rely on static, pre-defined encoders or require complex, non-scalable training procedures. Foundation model-based approaches struggle to generalize across diverse, multi-type graph structures typical of heterogeneous financial networks. The third challenge is to decouple feature generation from model training to enable feature selection without end-to-end retraining or reliance on fixed encoder sets (Section~\ref{sec:method}).
\end{itemize}

To address these challenges, we present \our (\underline{G}raph \underline{F}eature \underline{A}utomated \underline{S}election), a distributed graph feature selection system deployed in Alipay. Our design is based on the observation that decoupling non-parametric graph feature aggregation from downstream model training enables both scalability and interpretability. By generating explicit structural statistics rather than learned embeddings, \our achieves seamless integration with tabular learning models while maintaining native explainability.

In \our, graph features are constructed through multi-hop subgraph extraction and multi-scale aggregation without learned parameters, enabling CPU-based execution and good distributed scalability. The distributed feature selection module extends the Boruta algorithm with median-based aggregation across partitions, making the procedure less sensitive to outlier partitions arising from skewed class distributions. This design enables automated feature selection at scale while reducing the need for manual expert intervention.

We evaluate \our on eight public benchmarks and three industrial datasets~\footnote{The data used in this research does not involve any Personal Identifiable Information(PII) and were all processed by data abstraction and data encryption, and the researchers were unable to restore the original data. Sufficient data protection was carried out during the process of experiments to prevent the data leakage and the data was destroyed after the experiments were finished. The data is only used for academic research and sampled from the original data, therefore it does not represent any real business situation in Ant Financial Services Group.}. \our performs competitively on public benchmarks and shows strong results on industrial datasets. Deployed on large-scale transaction networks, \our processes millions of seed nodes daily across multiple risk control scenarios. The technical contributions are summarized as follows:

\begin{itemize}[leftmargin=*]
\item \textbf{A scalable graph feature generation module.} We propose a scalable feature generation module that constructs explicit, interpretable structural features via multi-hop subgraph extraction and multi-scale aggregation without learned parameters. This enables efficient processing of large-scale industrial graphs while producing TreeSHAP-compatible tabular features (Section~\ref{sec:method_generation}).

\item \textbf{A distributed Boruta-based feature selection module.} We extend the Boruta algorithm with median-based aggregation across partitions, enabling robust distributed feature selection for large-scale graph data. This approach provides resilience against outlier partitions while automatically identifying informative features without domain expertise (Section~\ref{sec:method_selection}).

\item \textbf{Industrial deployment and evaluation.} We demonstrate the practical effectiveness of \our through deployment in Alipay. Our system achieves over $10\times$ reduction in feature engineering cycle time while uncovering fraud patterns with substantially higher detection rates than expert-crafted baselines (Section~\ref{sec:experiments}).
\end{itemize}

\section{Background and Related Work}
\label{sec:background}

\subsection{Problem Formulation}
\label{sec:problem}

We reformulate graph representation learning by decoupling feature aggregation from end-to-end training.

Formally, let $G = (V, E)$ denote a graph with node set $V$ and edge set $E$. Traditional GNNs update node representations through parameterized neighborhood aggregation:
\begin{align}
\mathbf{h}_v^{(l+1)} = \phi^{(l)} \left( \mathbf{h}_v^{(l)} ; \bigoplus_{u \in \mathcal{N}(v)} \psi^{(l)}\left( \mathbf{h}_u^{(l)}, \mathbf{h}_v^{(l)}, \mathbf{e}_{uv} \right) \right),
\end{align}
where $\mathbf{h}_v^{(l)}$ denotes the embedding of node $v$ at layer $l$, $\mathbf{e}_{uv}$ denotes the edge from $u$ to $v$, $\mathcal{N}(v)$ is its neighborhood, $\psi^{(l)}(\cdot)$ is a parameterized feature aggregation function, $\bigoplus$ is a permutation-invariant aggregation operator, and $\phi^{(l)}(\cdot)$ is an update function.

Instead of learning parameterized aggregation functions, we perform \emph{non-parametric feature aggregation} through algorithmic graph transformations:
\begin{equation}
\mathbf{F} = \mathcal{A}_{\text{NP}}(G),
\end{equation}
where $\mathcal{A}_{\text{NP}}(\cdot)$ denotes a parameter-free operator that computes graph-level and node-level statistics without gradient-based optimization.
To enhance discriminative power and reduce redundancy, an \emph{automated feature selection} operator $\mathcal{S}(\cdot)$ is subsequently applied:
\begin{equation}
\mathbf{F}^\ast = \mathcal{S}(\mathbf{F}; \Phi),
\end{equation}
where $\Phi$ denotes the distributed selection configuration, including partition-wise importance estimation, cross-partition aggregation, and final tentative-feature ranking.
The selected features $\mathbf{F}^\ast$ are then utilized by a lightweight predictive model:
\begin{equation}
\hat{Y} = f_\theta(\mathbf{F}^\ast),
\end{equation}
Finally, a task-specific loss function $\mathcal{L}(Y, \hat{Y})$ is applied to optimize the predictive model. This framework supports efficient processing of industrial-scale graphs while preserving feature interpretability.

\subsection{Graph Feature Definition}
\label{sec:graph_feature}

Graph features capture structural characteristics through graph metrics and aggregation functions. Table~\ref{tab:graph_metrics} summarizes twelve metrics classified into three categories~\cite{gmetrics}: distance-based, connection-based, and spectral. We combine these with non-parametric aggregators~\cite{sgc,pmlp} to integrate neighbor features while maintaining interpretability.

\begin{table}[t]
\centering
\caption{Summary of graph metrics by category.}
\label{tab:graph_metrics}
\resizebox{\linewidth}{!}{
\begin{tabular}{llll}
\toprule
\textbf{Category} & \textbf{Distance} & \textbf{Connection} & \textbf{Spectral} \\
\midrule
\multirow{4}{*}{Metric}
& Hopcount & Degree & Algebraic connectivity \\
& Closeness & Entropy & Spectral radius \\
& Eccentricity & Assortativity & Spectral partitioning \\
& Diameter & Coreness & Principal eigenvector \\
\bottomrule
\end{tabular}}
\end{table}

\begin{figure}[t!]
  \centering
  \begin{minipage}[t]{0.43\linewidth}
    \centering
    \includegraphics[width=\linewidth]{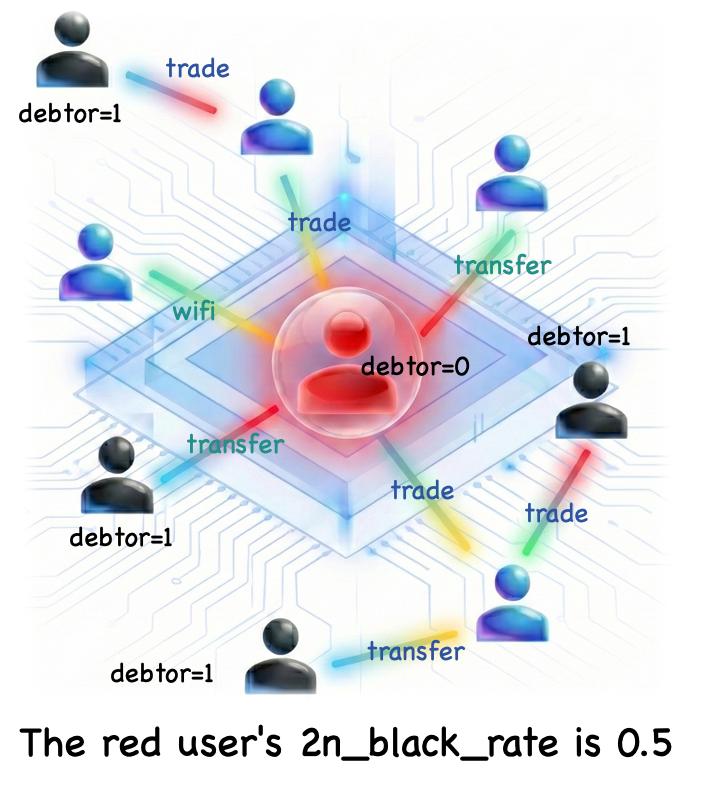}
    \caption{An example of graph-based features.}
    \label{fig:real_case}
  \end{minipage}
  \hfill
  \begin{minipage}[t]{0.48\linewidth}
    \centering
    \includegraphics[width=\linewidth]{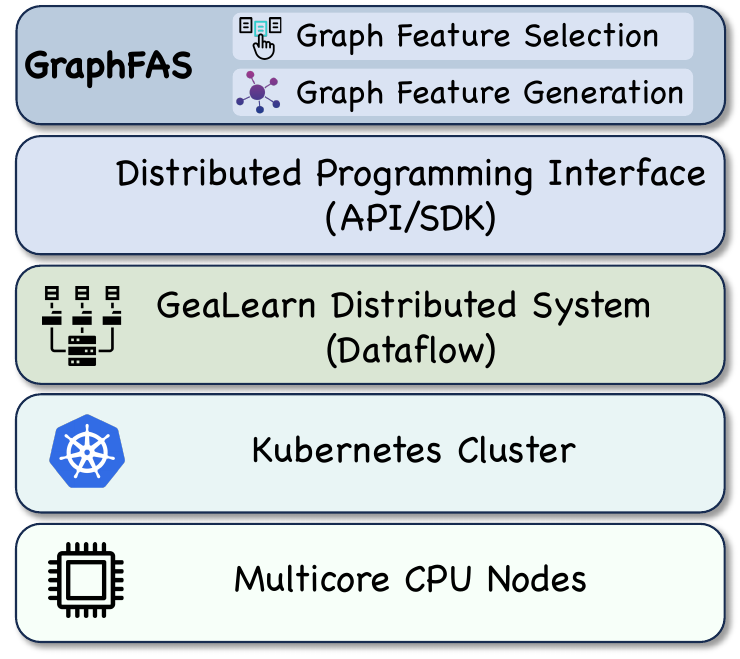}
    \caption{System architecture of GraphFAS.}
    \label{fig:graphfas_system}
  \end{minipage}
\end{figure}

\subsection{Related Work}
\label{sec:related_work}

\textbf{Feature Selection Methods.} Wrapper-based methods like Boruta~\cite{JSSv036i11} identify significant features by comparing them against randomly permuted shadow features. While effective~\cite{10.1145/3736761,MANIKANDAN2024101442}, existing implementations lack distributed support, hindering scalability for large-scale graph datasets. Filter methods (mutual information, statistical tests) are efficient but ignore feature interactions and structural dependencies.

\textbf{Graph Representation Learning.} Random walk methods (DeepWalk~\cite{DBLP:journals/corr/PerozziAS14}, Node2Vec~\cite{DBLP:journals/corr/GroverL16}) and spectral approaches produce node embeddings but lack interpretability and fail to incorporate node attributes. GNNs (GCN~\cite{gcn}, GAT~\cite{gat}) achieve strong performance through message passing but suffer from three critical limitations in industrial settings: (1) opaque embeddings violate regulatory transparency requirements; (2) end-to-end training introduces significant latency that conflicts with low-latency feature production requirements on large-scale networks; (3) latent representations lack semantic meaning for actionable analysis.

Post-hoc explainability methods (GNNExplainer~\cite{ying2019gnnexplainergeneratingexplanationsgraph}, PGExplainer~\cite{luo2020parameterizedexplainergraphneural}) generate soft masks highlighting important subgraphs. However, soft masks require thresholding for practical use, while hard masks are more applicable in industrial settings~\cite{amara2024graphframexsystematicevaluationexplainability}. These methods produce approximations rather than exact attributions and cannot directly output tabular features for downstream tasks~\cite{NIPS2017_6449f44a}. In contrast, our approach generates inherently interpretable structural statistics compatible with native TreeSHAP explainability.

\textbf{Graph-to-Tabular Methods.} Recent approaches (TAG~\cite{TAG}, G2T-FM~\cite{G2T-FM}, TabPFN-GN~\cite{TabPFN-GN}, GraphPFN~\cite{GraphPFN}) automate graph feature generation for tabular models. However, they rely on static pre-defined encoders or require complex training, limiting adaptability to diverse graph structures and scalability to industrial networks.

\subsection{Research Gap and Motivation}
\label{sec:gap}

Existing approaches exhibit three technical limitations that motivate our work. First, wrapper-based feature selection methods lack distributed adaptations---Boruta operates on standalone environments without partition-aware aggregation mechanisms, making it intractable for large-scale datasets. Second, current graph learning methods force a trade-off between performance and interpretability: GNNs produce opaque embeddings unsuitable for regulatory audit, while post-hoc explainers provide approximations rather than exact attributions. Third, automated graph-to-tabular pipelines rely on static pre-defined encoders that cannot adapt to diverse heterogeneous graph structures without complex retraining.  We leave such benchmarking to future work.

\our addresses these gaps through: (1) a distributed Boruta adaptation with median-based aggregation for robust feature selection at scale; (2) non-parametric graph feature generation producing inherently interpretable structural statistics compatible with native TreeSHAP explainability; (3) decoupled feature generation and automated selection enabling scalable deployment without end-to-end retraining.

\section{The GraphFAS Framework}
\label{sec:method}

\begin{figure}[t]
  \centering
  \includegraphics[width=\linewidth]{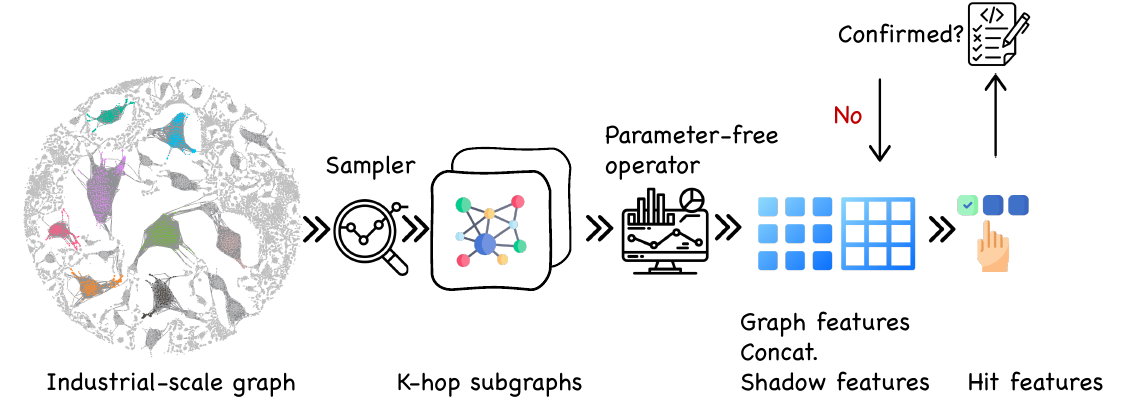}
  \caption{\our procedure overview.}
  \label{fig:graphfas}
\end{figure}

The calculation procedure for \our (Figure~\ref{fig:graphfas}) comprises two key design elements: (1) graph feature generation, which extracts $k$-hop subgraphs from seed nodes, computes graph metrics, and aggregates them to create candidate features; and (2) distributed feature selection, which employs a Boruta-based algorithm to filter out low-importance features.

\subsection{Graph Feature Generation}
\label{sec:method_generation}

The feature generation process involves three stages:

\mypara{K-hop Subgraph Extraction.}
For each node, ego-subgraphs are generated at different hop levels (1-hop, 2-hop, 3-hop) using neighborhood sampling, capturing localized structural patterns at varying depths.

\mypara{Feature Generation from Graph Metrics.}
Given a subgraph centered at target node $v$, we compute graph metric functions to extract feature values characterizing local structural properties (connectivity patterns, centrality, neighborhood composition). These metric-based features are concatenated with the aggregated graph feature vector, yielding an enhanced representation encoding both intrinsic attributes and structural context.

All generated features are inherently interpretable---e.g., \emph{average transaction frequency among 2-hop risky neighbors} or \emph{fraud concentration ratio within immediate neighborhood} (Figure~\ref{fig:real_case} illustrating an example of graph-based features highlighting suspect connectivity to known debtors. A node with 0.2 debtor ratio in 1-hop and 1.0 in 2-hop neighborhood suggests elevated risk even when the individual is not a debtor, demonstrating the capacity to identify latent risks through local topology)---enabling direct validation by domain experts. Unlike conventional pipelines requiring manual metric selection, \our automatically evaluates and ranks features from a large candidate pool, ensuring both predictive performance and scalability.

\mypara{Feature Aggregation.}
Multi-scale features are aggregated hierarchically: 1-hop aggregation (Mean/Max pooling of immediate neighbors), 2-hop aggregation (variance/skewness of secondary neighbors), and combined features (concatenation of raw features with aggregated features). We employ six complementary aggregation functions: Max/Min (extreme behaviors), Mean/Std (distributional properties), Sum (cumulative effects), and Count (structural density). Implementation hyperparameters are detailed in Table~\ref{tab:hyperparams}.

\subsection{Distributed Feature Selection}
\label{sec:method_selection}

\our employs a wrapper-style feature selection stage that iteratively evaluates feature importance against shuffled shadow features. For industrial-scale graphs, \our employs partition-based selection: the graph is divided into $w$ parts, each generating hybrid shadow features via random permutation (Algorithm~\ref{algo:graphfas_feature_selection}).

\begin{algorithm}[t!]
\caption{Feature Selection of \our}
\label{algo:graphfas_feature_selection}
\scriptsize
\KwIn{Dataset $D$, partitions $w$, max iterations $T$}
\KwOut{Confirmed features $F_{\text{final}}$}

$\{D_i\}_{i=1}^w \gets \text{Partition}(D, w)$\;
$\forall f_j \in F, \text{state}_j \gets \text{Tentative}$\;
Generate shadow features $\widetilde{F}_i$ on each worker $P_i$\;

\For{$t = 1$ \textbf{to} $T$}{
\tcp{Local importance computation}
\For{each partition $D_i$}{Compute $Z_j^{(t,i)}$ for $f_j \in F$, $\widetilde{Z}_k^{(t,i)}$ for $\widetilde{f}_k \in \widetilde{F}_i$\; \label{line:compute_importance_score}}
$\widetilde{Z}_{\text{max}}^{(t)} \gets \max_i \widetilde{Z}^{(t,i)}$; $\widetilde{Z}_{\text{min}}^{(t)} \gets \min_i \widetilde{Z}^{(t,i)}$\;
\tcp{Median aggregation across partitions}
\For{each feature $f_j \in F$ with $\text{state}_j = \text{Tentative}$}{
$\overline{Z}_j^{(t)} \gets \text{Median}(\{Z_j^{(t,i)}\}_{i=1}^w)$\;
\uIf{$\overline{Z}_j^{(t)} > \widetilde{Z}_{\text{max}}^{(t)}$}{$\text{state}_j \gets \text{Confirmed}$\;}
\uElseIf{$\overline{Z}_j^{(t)} < \widetilde{Z}_{\text{min}}^{(t)}$}{$\text{state}_j \gets \text{Rejected}$\;}}
\lIf{no Tentative features remain}{\textbf{break}}}

\tcp{Rank-based retention of tentative features}
$\text{Rank}(f_j) \gets \underset{t}{\mathrm{median}}\left(Z_j^{(t)} / \widetilde{Z}_{\text{max}}^{(t)}\right)$ for all $f_j$\;
$F_{\text{final}} \gets \{f_j \mid \text{state}_j = \text{Confirmed}\} \cup \text{Top-K}(\text{Rank})$\; \label{line:get_final_rank}
\KwOut{$F_{\text{final}}$}
\end{algorithm}

In each iteration, each partition calculates importance for candidate and shadow features (Line~\ref{line:compute_importance_score}). The median importance across partitions determines feature status: confirmed if exceeding all shadows, rejected if below all. After iterations, confirmed features plus top-$k$ ranked tentative features are selected (Line~\ref{line:get_final_rank}).

\mypara{Hyperparameter Settings.}
Key hyperparameters are summarized in Table~\ref{tab:hyperparams}: max iterations $T=100$, median aggregation, and top-$k=500$ fallback for tentative features.

\mypara{Connection to Explainability.}
The feature selection process naturally supports post-hoc explainability.
Since selected features are explicit structural statistics (e.g., "2-hop fraud neighbor ratio"), TreeSHAP can directly attribute predictions to human-interpretable graph patterns without additional approximation techniques required by GNN embeddings.
This straightforward integration facilitates model interpretation in compliance-oriented settings.

\mypara{Importance Score Computation.}
The importance score of features can be computed through   \emph{impurity-based metrics} (e.g., Gini coefficient) or \emph{Shapley value-based explanations}~\cite{wang2024feature}. LightGBM natively supports both approaches, with distinct computational characteristics for each.


Shapley values~\cite{shapley1951value} quantify the marginal contribution of each feature across all possible feature subsets. Exact computation is NP-hard with exponential complexity $O(2^{|F|})$, where $F$ denotes the set of all input features. TreeSHAP~\cite{DBLP:journals/corr/abs-1802-03888} leverages the internal structure of tree-based models to reduce complexity to polynomial time $O(T \cdot L \cdot |F|^2)$, where $T$ is the number of trees and $L$ is the maximum tree depth. By recursively traversing decision tree paths rather than enumerating all permutations, TreeSHAP enables practical application in domains requiring both transparency and computational efficiency.

In \our, we employ TreeSHAP for feature importance computation due to its scalability on industrial-scale graphs.

\mypara{Final Ranking.}
Features are ranked by stability score: median normalized importance relative to maximum shadow value. Let $Z_j^{(t)}$ denote the median importance score of feature $f_j$ at iteration $t$ aggregated across all partitions (i.e., $Z_j^{(t)} = \overline{Z}_j^{(t)}$ in Algorithm~\ref{algo:graphfas_feature_selection}), and $\widetilde{Z}^{(t)}$ denote the importance scores of shadow features at iteration $t$. The final rank is computed as:
\begin{equation}\text{Rank}(f_j) = \underset{t=1 \dots T}{\mathrm{median}} \left(\frac{Z_j^{(t)}}{\max(\widetilde{Z}^{(t)})}\right)
\end{equation}
Top-$k$ features are selected from tentative features by this rank.

\section{Distributed Implementation}
\our is implemented on \GeaLearning{}~\cite{liu2023graphthetadistributedgraphneural,tian2024graphrpm,jin2025graphgenadvancingdistributedsubgraph,wu2025distributedgraphneuralnetwork}, a distributed graph computing system employing a Manager-Worker architecture. The Manager initializes the cluster topology, coordinates distributed execution, and aggregates feature importance scores across workers. It instantiates a Driver module that encapsulates the algorithmic logic. Workers execute parallel graph operations with dynamic workload monitoring. This architecture decouples control from computation, enabling \our to process massive graph datasets with optimal resource utilization.

Figure~\ref{fig:graphfas_system} depicts the layered system architecture of \our. The top layer comprises the core GraphFAS components: feature generation and feature selection modules. These interface with the underlying distributed computing infrastructure through the GeaLearn distributed programming interface, which manages dataflow across a Kubernetes cluster deployed on multicore CPU nodes.

\subsection{Distributed~\our}

\begin{lstlisting}[language=C++, style=apistyle, caption={Implementation of GraphFAS}, captionpos=b, label=lst:graphfas_code, numberstyle=\tiny\color{gray}, numbersep=8pt]
class GraphFAS: public gealearn::Driver {
  void run(gealearn::DriverContext& context){
    // Load graph
    context.runProcedure("GraphFASGraphImport");
    // Feature generation
    context.runProcedure("GraphFASFeatureGeneration");
    features.initialize("tentative");
    // Feature selection
    for(int iter = 0; iter < max_iteration &&
        features.exist("tentative"); iter++){
        context.runProcedure("calculateImportance");
        context.allReduce(features);
        for(auto feature: features){
            if(feature.median > features.shadow.max)
                feature.set("confirmed");
            if(feature.median < features.shadow.min)
                feature.set("rejected");
        }
    }
    // Rank features
    context.runProcedure("GraphFASFeatureRanking");
  }
};
\end{lstlisting}

Our distributed implementation of \our adheres to the Manager-Worker paradigm of \GeaLearning{}. In this setup, the manager node is responsible for loading the graph and synchronizing the importance scores of features across all worker nodes. 

In alignment with the description provided in Algorithm~\ref{algo:graphfas_feature_selection}, \our utilizes the median value of a feature to update its state. To facilitate synchronization in a distributed environment, an additional \texttt{allReduce} function has been incorporated. This function ensures that the importance scores are consistently aggregated and updated across all nodes.
We show in detail how each stage of \our is implemented in the distributed environment:

\mypara{Graph Loading.} Given the raw dataset $D \in \mathbb{R}^{n \times m}$ with $n$ samples and $m$ features, we employ 1D Distributed Sample Transposition (1D-DST) to efficiently distribute the data across multiple worker processes. The dataset $D$ is partitioned into $w$ shards, denoted as $\{D_i\}_{i=1}^w$, where each shard $D_i$ is assigned to a worker process $\mathbf{P}_i$. Each worker maintains a submatrix $D_i \in \mathbb{R}^{n_i \times m}$ with $n_i \approx \frac{n}{w}$, ensuring that the computational load is balanced across the clusters through dynamic workload monitoring. 

\mypara{Feature Generation in Distributed Environment.} To incorporate subgraph features as candidate features, \our employs a distributed sampling strategy for efficient extraction and processing. The process starts with generating $k$-hop subgraphs for seed nodes, performed in parallel across multiple worker nodes. Each worker collects edges associated with its assigned seed nodes; if an edge $E = (v_1, v_2) \in E$ belongs to multiple seed nodes, it is replicated across workers to maintain subgraph completeness. After extraction, workers independently calculate subgraph features, such as averaging node features within each subgraph, ensuring efficient and consistent feature generation.

\mypara{Distributed Training and Feature Selection.} Each worker in \our trains its own importance scoring model using its partitioned data, running in parallel to maximize computational efficiency. After training, the manager node synchronizes the importance scores from all workers. It then aggregates this information to select features with higher importance. Median aggregation provides resilience against outlier partitions arising from skewed class distributions, as the median is less sensitive to extreme values compared to the mean.

\section{Experiments}
\label{sec:experiments}

\subsection{Experimental Setup}
\label{sec:experiment_setup}

\subsubsection{Datasets}
\label{sec:datasets}

We evaluate \our on eight public benchmarks from PyTorch Geometric and SNAP, and three industrial datasets from Alipay. Table~\ref{tab:datasets} provides statistics.

\textbf{Public datasets.} Public datasets include eight citation, co-purchasing, and social networks from PyG and SNAP.
Small-scale datasets use 60\%/20\%/20\% train/validation/test splits; large-scale datasets (Flickr, Reddit) use standard predefined splits~\cite{zenggraphsaint}.

\textbf{Industrial datasets.} Dataset1-1M (2.8M edges), Dataset2-5M (10.9M edges), and Dataset3-97M (285.4M edges) are transaction networks with multi-relational edges (chatting, financial cooperation, payment, trade) and authentic fraud labels. Nodes have 87-dimensional features. Due to severe class imbalance ($<$0.1\% positive samples), we use AUC-ROC as the primary evaluation metric.

\begin{table}[t]
    \centering
    \caption{Statistics of public and industrial datasets}
    \label{tab:datasets}
    \scalebox{0.8}{
    \begin{tabular}{llllll}
    \toprule
    \textbf{Datasets} & \textbf{Nodes} & \textbf{Edges} & \textbf{Features} & \textbf{Classes} & \textbf{Seeds}\\ \midrule
    Cora~\cite{cora} & 2,708 & 5,429 & 1,433 & 7 & 2,708 \\
    CiteSeer~\cite{citeseer} & 3,186 & 4,277 & 3,703 & 6 & 3,186 \\
    PubMed~\cite{pubmed} & 19,717 & 44,338 & 500 & 3 & 19,717 \\
    DBLP~\cite{cora} & 17,716 & 105,734 & 1,639 & 4 & 17,716 \\
    Computers~\cite{shchur2018pitfalls} & 13,752 & 491,722 & 767 & 10 & 13,752 \\
    Photo~\cite{shchur2018pitfalls} & 7,650 & 238,162 & 745 & 8 & 7,650 \\
    Flickr~\cite{zenggraphsaint} & 89,250 & 899,756 & 500 & 7 & 89,250 \\
    Reddit~\cite{zenggraphsaint} & 232,965 & 23,213,838 & 602 & 41 & 232,965 \\ \midrule
    Dataset1-1M & 1,463,690 & 2,828,041 & 87 & 2 & 25K \\
    Dataset2-5M & 5,629,431 & 10,919,773 & 87 & 2 & 70K \\
    Dataset3-97M & 97,262,426 & 285,366,878 & 87 & 2 & 3.5M \\
    \bottomrule
    \end{tabular}}
\end{table}

\subsubsection{Implementation Settings}
\label{sec:settings}

Public dataset experiments use a 64-core Intel Xeon E5-2682 v4 CPU with 256GB RAM. Industrial experiments use Kubernetes production clusters. Evaluation metrics: Accuracy (small public datasets), Micro-F1 (large public datasets), AUC-ROC (industrial datasets). Hyperparameters are summarized in Table~\ref{tab:hyperparams}.

\begin{table}[t]
\centering
\caption{Hyperparameter settings of \our.}
\label{tab:hyperparams}
\resizebox{\linewidth}{!}{
\begin{tabular}{llp{5cm}}
\toprule
\textbf{Component} & \textbf{Parameter} & \textbf{Value / Setting} \\
\midrule
\multirow{3}{*}{Subgraph Extraction}
 & Max hop distance & 3 \\
 & Sampling budget & 10,000 nodes per seed \\
 & Neighborhood strategy & Random sampling \\
\midrule
\multirow{3}{*}{Feature Generation}
 & Graph metrics & 12 metrics (Table~\ref{tab:graph_metrics}) \\
 & Aggregation functions & mean, max, min, std, sum, count \\
 & Candidate pool size & $\sim$500 features \\
\midrule
\multirow{4}{*}{Feature Selection}
 & Max iterations ($T$) & 100 \\
 & Aggregation across partitions & Median \\
 & Shadow feature generation & Random permutation per partition \\
 & Final selection & Confirmed + top-500 tentative \\
\midrule
\multirow{5}{*}{Downstream Model}
 & Model & LightGBM \\
 & Learning rate & 0.05 \\
 & Num leaves & 31 \\
 & Feature fraction & 0.8 \\
 & Early stopping & 50 rounds patience \\
\bottomrule
\end{tabular}}
\end{table}

\subsection{Public Graph Benchmarks}
\label{sec:public}

\subsubsection{Baselines and Setup}
\label{sec:baselines_public}

We compare \our with twelve baselines: (1) \textbf{Traditional ML}: LR~\cite{logistic}, LightGBM~\cite{NIPS2017_6449f44a} and MLP; (2) \textbf{Deep graph learning}: GCN~\cite{gcn}, GraphSAGE~\cite{sage}, GAT~\cite{gat}, SGC~\cite{sgc}, GIPA~\cite{Zheng2021GIPAGI,li2023gipa}; (3) \textbf{Feature selection}: PCA, RFE, F-test and MI. All methods use grid search.

\subsubsection{RQ1: Comparison with GNN Methods} Table~\ref{tab:public} presents node classification results (mean±std over five runs).

GNNs like GCN outperform traditional methods (LR, LGBM) by over 10\% on Cora and 7\% on DBLP, indicating that inherent structural information is highly effective. \our harnesses this information via its Graph Feature Generation Module while employing automated Feature Selection to identify the most informative features and mitigate overfitting. \our achieves competitive results with SOTA GNNs on smaller datasets 
and over 1\% absolute improvement on larger datasets (DBLP, Flickr), demonstrating strong suitability for large-scale graphs.

\begin{table*}[t!]
    \centering
    \caption{Node classification accuracy on public benchmarks (\%).}
    \vspace{0.3em}
    \label{tab:public}
    \begin{tabular}{lllllllll}
    \toprule
    & \textbf{Cora} & \textbf{CiteSeer} & \textbf{PubMed} & \textbf{DBLP} & \textbf{Computers} & \textbf{Photo} & \textbf{Flickr} & \textbf{Reddit}\\
    \midrule
    LR & 76.34±1.33 & 71.41±0.88 & 87.55±0.57 & 75.10±0.66 & 84.14±0.29 & 92.03±0.27 & 46.62±0.08 & 52.41±0.02 \\
    LGBM & 76.57±1.31 & 71.92±1.72 & 90.86±0.34 & 75.05±0.46 & 86.38±0.28 & 92.88±0.50 & 46.92±0.12 & 70.40±0.03 \\
    MLP & 86.94±0.99 & 71.68±1.81 & 88.23±0.28 & 75.25±0.55 & 85.62±0.45 & 92.16±0.58 & 44.16±0.48 & 57.33±0.37 \\
    GCN & 88.15±1.09 & 76.61±0.49 & 89.16±0.61 & 82.17±0.59 & 90.64±0.76 & 93.45±0.78 & 53.13±0.51 & 92.21±0.20 \\
    SAGE & \textbf{88.41±1.24} & \textbf{77.08±0.75} & 89.39±0.41 & 83.94±0.29 & 91.44±0.22 & 95.59±0.34 & 53.10±0.65 & 93.11±0.21 \\
    GAT & 88.30±0.55 & 76.76±0.86 & 88.11±0.27 & 83.79±0.51 & 91.51±0.53 & 95.11±0.62 & 53.52±1.10 & 92.60±0.17 \\
    SGC & 87.49±0.82 & 75.53±0.70 & 87.06±0.30 & 83.19±0.49 & 91.01±1.01 & 93.49±0.28 & 51.13±0.11 & 91.74±0.14 \\
    GIPA & 86.75±1.26 & 72.85±1.61 & 89.12±0.69 & 84.13±0.60 & 91.57±0.88 & 95.14±0.50 & 53.73±0.93 & \textbf{95.91±0.25} \\
    \midrule
    PCA & 67.82±1.20 & 68.11±1.35 & 85.12±0.74 & 74.55±0.36 & 85.37±0.57 & 91.28±0.42 & 46.25±0.06 & 64.51±0.03 \\
    RFE & 74.17±1.44 & 70.03±0.68 & 90.12±0.30 & 74.17±0.25 & 86.88±0.26 & 92.18±0.50 & 46.09±0.05 & 66.32±0.02 \\
    F-test & 72.47±1.28 & 71.02±1.43 & 89.93±0.36 & 74.95±0.40 & 85.82±0.58 & 91.62±0.52 & 46.02±0.09 & 65.70±0.02 \\
    MI & 71.51±1.14 & 66.16±1.54 & 89.67±0.25 & 70.28±0.41 & 85.99±0.78 & 92.03±0.47 & 45.96±0.08 & 64.99±0.03 \\
    \midrule
    Ours & 88.27±1.15 & 76.19±1.02 & \textbf{92.09±0.30} & \textbf{85.02±0.49} & \textbf{91.54±0.56} & \textbf{95.78±0.38} & \textbf{54.56±0.06} & 95.82±0.03 \\
    \bottomrule
    \end{tabular}
\end{table*}

\subsubsection{RQ2: Comparison with Feature Selection Methods} Feature selection methods (MI, F-test) generally underperform compared to traditional ML due to information loss. For instance, RFE falls behind LGBM by approximately 2\% and 1\% on Cora and DBLP, indicating that aggressive feature pruning harms predictive accuracy. \our overcomes this by automatically generating and selecting informative graph features, achieving 10-20\% absolute improvements over feature selection baselines.

\subsection{Industrial Case Studies}
\label{sec:industrial}

\subsubsection{Baselines}
\label{sec:baselines_industrial}

We select GIPA~\cite{li2023gipa} as a representative industrial-strength baseline, given its demonstrated superiority over standard GNNs on public benchmarks and its deployment in production environments. Standard GNNs (e.g., GCN, GraphSAGE) consistently underperform GIPA on industrial datasets~\cite{li2023gipa} and are excluded to focus on competitive baselines.

\subsubsection{RQ3: Effectiveness on Fraud Detection}

\begin{table}[t]
    \centering
    \caption{Fraud detection performance on industrial datasets (AUC-ROC \%).}
    \label{tab:industrial}
    \begin{tabular}{llll}
    \toprule
    & \textbf{Dataset1} & \textbf{Dataset2} & \textbf{Dataset3}\\ \midrule
    GIPA& 80.12±1.24 & 98.03±0.56 & 95.21±0.88 \\ \midrule
    \our & \textbf{86.26±0.83} & \textbf{98.80±0.31} & \textbf{99.90±0.12}\\
    \bottomrule
    \end{tabular}
    \vspace{0.7em}
\end{table}

Table~\ref{tab:industrial} shows \our consistently outperforms GIPA on all three industrial datasets. The most substantial gain is on Dataset1 (2.82M edges): 6.14\% absolute AUC improvement (86.26\% vs. 80.12\%). On Dataset3 (285.37M edges), \our achieves very strong discrimination (99.90\% AUC-ROC) vs. GIPA's 95.21\%.

Three key insights emerge: (1) \our effectively captures discriminative patterns from multi-relational graphs with four distinct edge types; (2) it maintains strong performance under severe class imbalance ($<$0.1\% positive samples), demonstrating the automated Boruta-based mechanism's effectiveness; (3) consistent improvements across two orders of magnitude in graph size ($\sim$3M to $\sim$285M edges) suggest good distributed scalability (Section~\ref{sec:efficiency}).

\subsection{Efficiency Analysis and Ablation Study}
\label{sec:efficiency}

\subsubsection{RQ4: Scalability} We evaluate scalability on Dataset3-97M across 16, 32, and 64 containers (8 cores per container, five runs averaged). Execution times are 3866 seconds (16 nodes, 32GB per node, denoted 16N-32G), 2014 seconds (32N-16G), and 1075 seconds (64N-8G). Scaling from 16 to 32 containers reduces runtime by 47.9\% (3866 seconds $\rightarrow$ 2014 seconds), yielding 1.92$\times$ speedup; scaling from 32 to 64 containers achieves 1.87$\times$ speedup (2014 seconds $\rightarrow$ 1075 seconds). The overall speedup from 16 to 64 containers reaches 3.6$\times$, with sublinear scaling indicating growing communication and synchronization overheads at higher scales.

\subsubsection{RQ5: Sensitivity Analysis} 
\begin{figure}[t]
    \centering
    \includegraphics[width=0.85\linewidth]{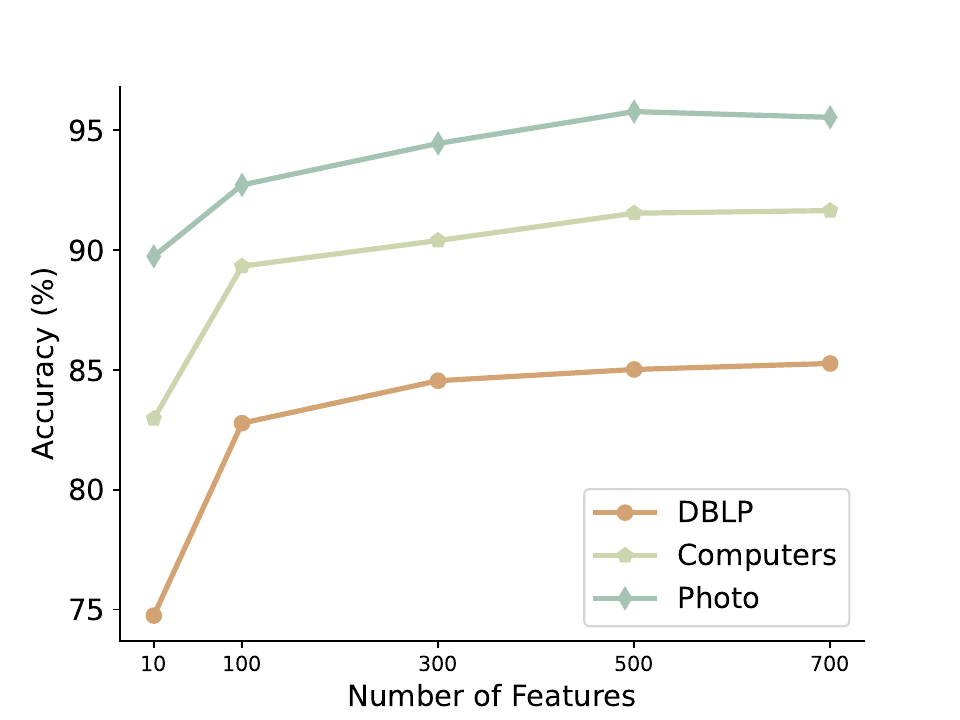}
    \caption{Sensitivity analysis on the number of features}
    \label{fig:sensitivity}
    \vspace{1em}
\end{figure}

Figure~\ref{fig:sensitivity} shows that selecting too few features leads to poor performance due to information loss. Performance improves with more features, stabilizing around 500. Beyond this threshold, adding features introduces noise (e.g., Photo at 700 features). We set the feature count to 500 across all datasets.

\subsubsection{RQ6: Module Effectiveness} 
\begin{figure}[t]
    \centering
    \includegraphics[width=\linewidth]{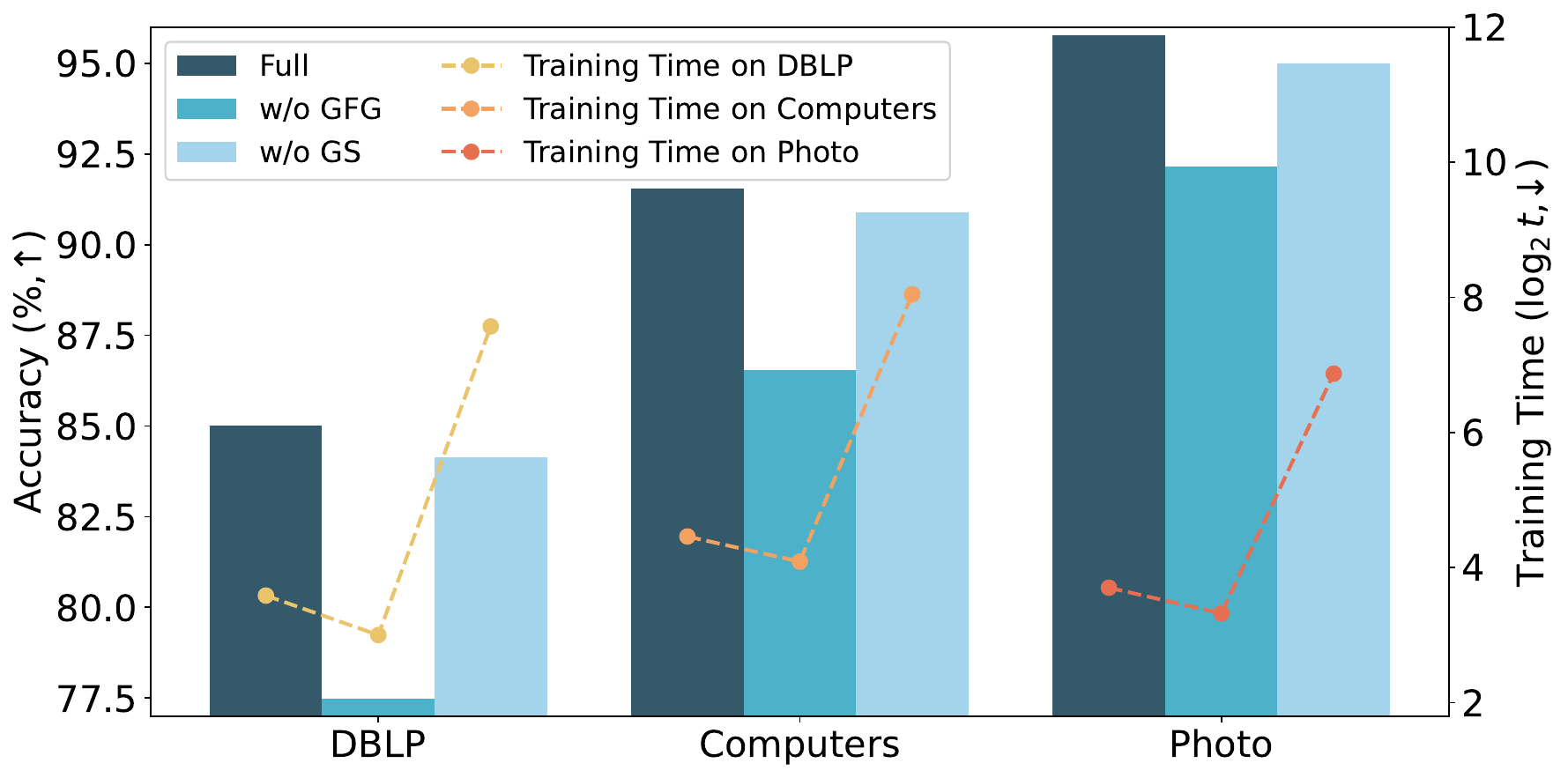}
    \caption{Ablation study across different datasets. Left y-axis: accuracy; right y-axis: $\log_2$(training time).}
    \label{fig:ablation}
    \vspace{0.7em}
\end{figure}

Figure~\ref{fig:ablation} presents the ablation study (training time measured on a single 64-core Intel Xeon E5-2682 v4 CPU). ``w/o GFG'' removes the Graph Feature Generation; ``w/o FS'' removes the Feature Selection. The full model provides the best overall trade-off between performance and runtime.

The full model significantly outperforms ``w/o GFG'' across all datasets, underscoring the critical importance of graph features. On Computers, \our (91.54\%) achieves comparable accuracy to ``w/o FS'' (90.89\%) but with over 10$\times$ speedup, demonstrating the FS module's effectiveness in eliminating irrelevant features without sacrificing predictive performance.

\section{Deployment and Industrial Impact}
\label{sec:deploy_impact}

\begin{figure}[t]
  \centering
  \includegraphics[width=1.\linewidth]{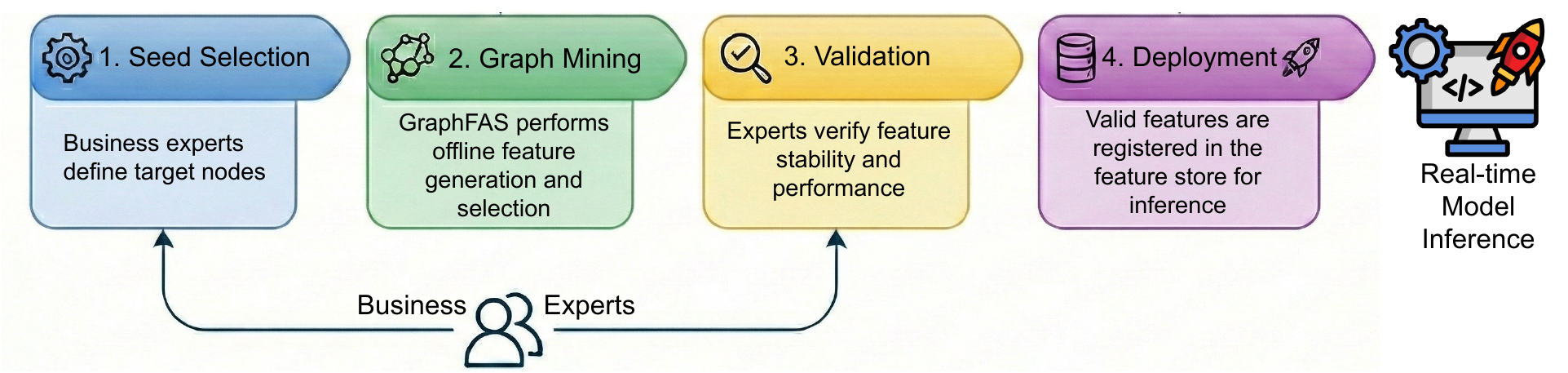}
  \caption{Operational workflow: from seed selection to production deployment}
  \Description{The process from an algorithm perspective.}
  \label{fig:production}
  \vspace{0.7em}
\end{figure}

\our has been deployed in production for over two years, handling millions of seed nodes daily across multiple risk control scenarios. The operational workflow (Figure~\ref{fig:production}) comprises: (1) \textbf{Seed Selection}---experts define target nodes; (2) \textbf{Graph Mining}---\our performs offline feature generation and selection; (3) \textbf{Validation}---experts verify feature stability; (4) \textbf{Deployment}---features are registered for real-time inference.

In a representative cash-out fraud detection scenario, \our~utilized credit relations and fraud model scores to identify high-risk graph patterns. Quantitatively, the identified features achieved a \textbf{tenfold lift} in uncovering latent fraud groups compared to baseline methods. Furthermore, by automating the discovery process, \our~ reduced the feature engineering cycle by \textbf{over 10$\times$} in our deployment compared to traditional manual assessment.

\section{Conclusion}

We present \our, a distributed graph feature selection system that provides a practical alternative to end-to-end GNN pipelines under industrial constraints for interpretability and scalability. By combining non-parametric graph feature generation with a distributed Boruta-style selection using median aggregation across partitions, \our~produces explicit, interpretable structural features compatible with native TreeSHAP explainability. Deployed at Alipay for two years, processing millions of seed nodes daily, it outperforms GNN-based approaches while achieving order-of-magnitude efficiency gains over manual feature engineering. Decoupling feature generation from model training sacrifices some representational capacity in exchange for practical benefits, including CPU-only execution with scalable distributed processing, audit-compliant structural statistics, and flexible downstream model updates without regenerating features.




\section{GenAI Usage Disclosure}
During the preparation of this work, we used Claude Code to assist with code development and manuscript writing. Specifically, the AI tool was utilized to generate boilerplate code, assist with implementation details, draft and polish text, and improve overall language clarity. All AI-generated content was thoroughly reviewed, verified, and refined by the authors. We assume full responsibility for the correctness of the code, the accuracy of the scientific claims, and the ultimate integrity of this work. The core research ideas, experimental design, data analysis, and scientific conclusions were entirely conceived and executed by our human authors.
%
\bibliographystyle{ACM-Reference-Format}
\balance
\bibliography{sample}  

\end{document}